\documentstyle{article}

\newif\ifNFSS   
\NFSSfalse

\input{psfig}

\newlength\titlebox 
 \twocolumn 

\makeatletter
\def\addcontentsline#1#2#3{}

\def\maketitle{\par
 \begingroup
   \def\thefootnote{\fnsymbol{footnote}}
   \def\@makefnmark{\hbox to 0pt{$^{\@thefnmark}$\hss}}
   \twocolumn[\@maketitle] \@thanks
 \endgroup
 \setcounter{footnote}{0}
 \let\maketitle\relax \let\@maketitle\relax
 \gdef\@thanks{}\gdef\@author{}\gdef\@title{}\let\thanks\relax}
\def\@maketitle{\vbox to \titlebox{\hsize\textwidth
 \linewidth\hsize \vskip 0.625in minus 0.125in \centering
 {\huge\bf \@title \par} \vskip 0.2in plus 1fil minus 0.1in
 {\def\and{\unskip\enspace{\rm and}\enspace}%
  \def\And{\end{tabular}\hss \egroup \hskip 1in plus 2fil
           \hbox to 0pt\bgroup\hss \begin{tabular}[t]{c}\Large\bf}%
  \def\AND{\end{tabular}\hss\egroup \hfil\hfil\egroup
	  \vskip 0.25in plus 1fil minus 0.125in
	   \hbox to \linewidth\bgroup\Large \hfil\hfil
 	     \hbox to 0pt\bgroup\hss \begin{tabular}[t]{c}\Large\bf}
  \hbox to \linewidth\bgroup\Large \hfil\hfil
    \hbox to 0pt\bgroup\hss \begin{tabular}[t]{c}\Large\bf\@author
			    \end{tabular}\hss\egroup
    \hfil\hfil\egroup}
  \vskip 0.3in plus 2fil minus 0.1in
}}

\def\section{\@startsection {section}{1}{\z@}{-2.0ex plus
    -0.5ex minus -.2ex}{3pt plus 2pt minus 1pt}{\Large\bf\raggedright%
}}
\def\subsection{\@startsection{subsection}{2}{\z@}{-2.0ex plus
    -0.5ex minus -.2ex}{3pt plus 2pt minus 1pt}{\large\bf\raggedright}}
\def\subsubsection{\@startsection{subparagraph}{3}{\z@}{-6pt plus
   2pt minus 1pt}{-1em}{\normalsize\bf}}

\skip\footins 9pt plus 4pt minus 2pt
\def\footnoterule{\kern-3pt \hrule width 5pc \kern 2.6pt }
\labelwidth\leftmargini\advance\labelwidth-\labelsep \labelsep 5pt

\def\@listi{\leftmargin\leftmargini}
\def\@listii{\leftmargin\leftmarginii
   \labelwidth\leftmarginii\advance\labelwidth-\labelsep
   \topsep 2pt plus 1pt minus 0.5pt
   \parsep 1pt plus 0.5pt minus 0.5pt
   \itemsep \parsep}
\def\@listiii{\leftmargin\leftmarginiii
    \labelwidth\leftmarginiii\advance\labelwidth-\labelsep
    \topsep 1pt plus 0.5pt minus 0.5pt
    \parsep \z@ \partopsep 0.5pt plus 0pt minus 0.5pt
    \itemsep \topsep}
\def\@listiv{\leftmargin\leftmarginiv
     \labelwidth\leftmarginiv\advance\labelwidth-\labelsep}
\def\@listv{\leftmargin\leftmarginv
     \labelwidth\leftmarginv\advance\labelwidth-\labelsep}
\def\@listvi{\leftmargin\leftmarginvi
     \labelwidth\leftmarginvi\advance\labelwidth-\labelsep}

\belowdisplayskip \abovedisplayskip
\def\@normalsize{\@setsize\normalsize{11pt}\xpt\@xpt}
\def\small{\@setsize\small{10pt}\ixpt\@ixpt}
\def\footnotesize{\@setsize\footnotesize{10pt}\ixpt\@ixpt}
\def\scriptsize{\@setsize\scriptsize{8pt}\viipt\@viipt}
\def\tiny{\@setsize\tiny{7pt}\vipt\@vipt}
\def\large{\@setsize\large{12pt}\xipt\@xipt}
\def\Large{\@setsize\Large{14pt}\xiipt\@xiipt}
\def\LARGE{\@setsize\LARGE{16pt}\xivpt\@xivpt}
\def\huge{\@setsize\huge{20pt}\xviipt\@xviipt}
\def\Huge{\@setsize\Huge{23pt}\xxpt\@xxpt}
\makeatother

\newcommand{\bkslf}{$ \backslash $}		

\newcommand{\RULE}[2]		                
{\begin{tabular}{c}
 \( {#1} \) \\ \hline \( {#2} \)
\end{tabular}}

\newcommand{\reftable}[1]{table~\ref{#1}}	
\newcommand{\reffig}[1]{figure~\ref{#1}}	

\newcommand{\refsec}[1]{section~\ref{#1}}

\newcommand{\comment}[1]{}
\newcommand{\GoesTo}{\(\longrightarrow\)}
\newcommand{\forw}{\verb+>+}
\newcommand{\back}{\verb+<+}
\newcommand{\up}{\(\mid\)}
\newcommand{\cB}{{\bf B}} 

\newcommand{\SUB}[1]{ \( \! \!_{ {#1}}\)}

\newcommand{\SUP}[1]{ \( \! \!^{ {#1}}\)}

\newcommand{\pf}[1]{\noindent{\bf proof }  {#1} \hfill~$\Box$}
\newcommand{\implies}{\: \supset \:}
\newcommand{\And}{\: \wedge \:}
\newcommand{\arr}{-\!\!\!\!\rightarrow}   			
\newcommand{\larr}{\leftarrow\!\!\!\!-}				
\newcommand{\darr}{-\!\!\!\!\rightarrow\!\!\!\!\!\rightarrow}   
\newcommand{\invdarr}{\leftarrow\!\!\!\!\!\leftarrow\!\!\!\! -} 
\newcommand{\drpl}{\invdarr\!\!\!\!\!\!\darr}   		

\newtheorem{defi}{Definition}
\newtheorem{LEMMA}{Lemma}
\newtheorem{COROLLARY}{Corollary}
\newtheorem{thm}{Theorem}

\newcommand{\lemma}[1]{\begin{LEMMA} {\rm {#1}} \end{LEMMA}}

\newcommand{\theorem}[1]{\begin{thm} {\rm {#1}} \end{thm}}
\newcommand{\kases}{    
   \vspace{-3.2mm}
   \begin{list}
 {???}
 {\setlength{\leftmargin}{4.1mm}
  \setlength{\labelwidth}{2mm}}
  \setlength{\parsep}{1mm}
  \setlength{\itemsep}{0.01mm}
  \setlength{\topsep}{0.01mm}
  \setlength{\parskip}{0mm}
  \setlength{\parindent}{0mm}}
\newcommand{\Endkases}{\end{list}\vspace{-3.2mm}}
\newcommand{\kase}[1]{\item[case {#1}:\ ]}

\newcommand{\State}[5]
{\mbox{\vbox{state #1:\\
\parbox[t]{2em}{\verb+ + B:} \parbox[t]{\ExampleWidth}{#2}\\
\parbox[t]{2em}{\verb+ + S:} \parbox[t]{\ExampleWidth}{#3}\\
\parbox[t]{2em}{\verb+ + I:} \parbox[t]{\ExampleWidth}{#4}\\
\parbox[t]{2em}{\verb+ + P:} \parbox[t]{\ExampleWidth}{#5}
}}}

\newcommand{\?}{$ \backslash $}

\newlength{\DerivSep}
\newlength{\DefaultLineThickness}
\newcommand{\NLN}[1]{\LN[0pt]{\rule{0pt}{1ex}}{#1}}

\makeatletter

\newsavebox{\LNbox}
\newlength{\LNlength}

\def\LN{\@ifnextchar[{\LNz}{\LNz[\DefaultLineThickness]}}

\def\LNz[#1]#2#3{{%
\savebox{\LNbox}{%
\setlength{\arrayrulewidth}{#1}%
\begin{tabular}{@{\hspace{.1em}}c@{\hspace{.1em}}}{#2}\\[.2ex]%
\hline\raisebox{-.2ex}{#3}\end{tabular}}%
\settowidth{\LNlength}{\usebox{\LNbox}}%
\makebox[\LNlength]%
{\begin{minipage}[t]{\LNlength}\usebox{\LNbox}\end{minipage}}}}

\newsavebox{\UNb}
\newlength{\UNl}

\def\UN{\@ifnextchar[{\UNz}{\UNz[\DefaultLineThickness]}}

\def\UNz[#1]#2#3#4{{
\savebox{\UNb}{#3}\settowidth{\UNl}{\usebox{\UNb}}%
\begin{minipage}[t]{\UNl}{\begin{center}\makebox[\UNl]{\usebox{\UNb}}\\
\makebox[\UNl][l]{%
\raisebox{-1.0ex}[.1ex][-.2ex]{\rule[0.65ex]{\UNl}{#1}\hspace{.1em}{#4}}}\\
{#2}\end{center}}\end{minipage}}}

\newsavebox{\BNba}
\newsavebox{\BNbb}
\newlength{\BNla}
\newlength{\BNlb}
\newlength{\BNl}

\def\BN{\@ifnextchar[{\BNz}{\BNz[\DefaultLineThickness]}}

\def\BNz[#1]#2#3#4#5{{
\savebox{\BNba}{#3}\settowidth{\BNla}{\usebox{\BNba}}%
\savebox{\BNbb}{#4}\settowidth{\BNlb}{\usebox{\BNbb}}\setlength{\BNl}{\BNlb}%
\addtolength{\BNl}{\BNla}%
\addtolength{\BNl}{\DerivSep}%
\begin{minipage}[t]{\BNl}{\begin{center}\makebox[\BNl]{%
\usebox{\BNba}%
\hspace*{\DerivSep}%
\usebox{\BNbb}}\\
\makebox[\BNl][l]{%
\raisebox{-1.0ex}[.1ex][-.2ex]{\rule[0.65ex]{\BNl}{#1}\hspace{.1em}{#5}}}\\
{#2}\end{center}}\end{minipage}}}

\newsavebox{\TNba}
\newsavebox{\TNbb}
\newsavebox{\TNbc}
\newlength{\TNla}
\newlength{\TNlb}
\newlength{\TNlc}
\newlength{\TNl}

\def\TN{\@ifnextchar[{\TNz}{\TNz[\DefaultLineThickness]}}

\def\TNz[#1]#2#3#4#5#6{
\savebox{\TNba}{#3}\settowidth{\TNla}{\usebox{\TNba}}%
\savebox{\TNbb}{#4}\settowidth{\TNlb}{\usebox{\TNbb}}%
\savebox{\TNbc}{#5}\settowidth{\TNlc}{\usebox{\TNbc}}\setlength{\TNl}{\TNlc}%
\addtolength{\TNl}{\TNla}%
\addtolength{\TNl}{\TNlb}%
\addtolength{\TNl}{2\DerivSep}%
\begin{minipage}[t]{\TNl}{\begin{center}\makebox[\TNl]{%
\makebox[\TNla]{\usebox{\TNba}}%
\hspace*{\DerivSep}\makebox[\TNlb]{\usebox{\TNbb}}%
\hspace*{\DerivSep}\makebox[\TNlc]{\usebox{\TNbc}}}\\
\makebox[\TNl][l]{%
\raisebox{-1.0ex}[.1ex][-.2ex]{\rule[0.65ex]{\TNl}{#1}\hspace{.1em}{#6}}}\\
{#2}\end{center}}\end{minipage}}

\newsavebox{\PivotBox}
\newlength{\PivotLength}
\newcommand{\SetPivot}[1]{%
\savebox{\PivotBox}{#1}%
\settowidth{\PivotLength}{\usebox{\PivotBox}}}

\newcommand{\UsePivot}{\usebox{\PivotBox}}

\def\UNp{\@ifnextchar[{\UNpz}{\UNpz[\DefaultLineThickness]}}

\def\UNpz[#1]#2#3#4{{
\savebox{\UNb}{#3}\settowidth{\UNl}{\usebox{\UNb}}%
\begin{minipage}[t]{\UNl}{%
\begin{center}\makebox[\UNl]{\usebox{\UNb}}\\
\makebox[\UNl]{%
\rule{\PivotLength}{0pt}%
\rule{\DerivSep}{0pt}%
\addtolength{\UNl}{-\PivotLength}%
\addtolength{\UNl}{-\DerivSep}%
\makebox[\UNl]{%
\begin{minipage}{\UNl}%
\begin{center}%
\makebox[\UNl][l]{%
\raisebox{-1.3ex}[.1ex][-.2ex]{\rule[0.55ex]{\UNl}{#1}\hspace{.1em}{#4}}}\\
\addtolength{\UNl}{\PivotLength}%
\addtolength{\UNl}{\DerivSep}%
{#2}%
\end{center}%
\end{minipage}}}%
\end{center}}%
\end{minipage}}}

\def\BNp{\@ifnextchar[{\BNpz}{\BNpz[\DefaultLineThickness]}}

\def\BNpz[#1]#2#3#4#5{{
\savebox{\BNba}{#3}\settowidth{\BNla}{\usebox{\BNba}}%
\savebox{\BNbb}{#4}\settowidth{\BNlb}{\usebox{\BNbb}}\setlength{\BNl}{\BNlb}%
\addtolength{\BNl}{\BNla}%
\addtolength{\BNl}{\DerivSep}%
\begin{minipage}[t]{\BNl}{%
\begin{center}\makebox[\BNl]{%
\makebox[\BNla]{\usebox{\BNba}}\hspace*{\DerivSep}\makebox[\BNlb]
	{\usebox{\BNbb}}}\
\makebox[\BNl]{%
\rule{\PivotLength}{0pt}%
\rule{\DerivSep}{0pt}%
\addtolength{\BNl}{-\PivotLength}%
\addtolength{\BNl}{-\DerivSep}%
\makebox[\BNl]{%
\begin{minipage}{\BNl}%
\begin{center}%
\makebox[\BNl][l]{%
\raisebox{-1.3ex}[.1ex][-.2ex]{\rule[0.55ex]{\BNl}{#1}\hspace{.1em}{#5}}}\\
\addtolength{\BNl}{\PivotLength}%
\addtolength{\BNl}{\DerivSep}%
{#2}%
\end{center}%
\end{minipage}}}%
\end{center}}%
\end{minipage}}}

\def\TNp{\@ifnextchar[{\TNpz}{\TNpz[\DefaultLineThickness]}}

\def\TNpz[#1]#2#3#4#5#6{{
\savebox{\TNba}{#3}\settowidth{\TNla}{\usebox{\TNba}}%
\savebox{\TNbb}{#4}\settowidth{\TNlb}{\usebox{\TNbb}}%
\savebox{\TNbc}{#5}\settowidth{\TNlc}{\usebox{\TNbc}}\setlength{\TNl}{\TNlc}%
\addtolength{\TNl}{\TNla}%
\addtolength{\TNl}{\TNlb}%
\addtolength{\TNl}{\DerivSep}%
\addtolength{\TNl}{\DerivSep}%
\begin{minipage}[t]{\TNl}{%
\begin{center}\makebox[\TNl]{%
\makebox[\TNla]{\usebox{\TNba}}\hspace*{\DerivSep}%
\makebox[\TNlb]{\usebox{\TNbb}}\hspace{\DerivSep}%
\makebox[\TNlc]{\usebox{\TNbc}}}\\
\makebox[\TNl]{%
\rule{\PivotLength}{0pt}%
\rule{\DerivSep}{0pt}%
\addtolength{\TNl}{-\PivotLength}%
\addtolength{\TNl}{-\DerivSep}%
\makebox[\TNl]{%
\begin{minipage}{\TNl}%
\begin{center}%
\makebox[\TNl][l]{\raisebox{-1.3ex}[.1ex][-.2ex]{\rule[0.55ex]
	{\TNl}{#1}\hspace{.1em}{#6}}}\\
\addtolength{\TNl}{\PivotLength}%
\addtolength{\TNl}{\DerivSep}%
{#2}%
\end{center}%
\end{minipage}}}%
\end{center}}%
\end{minipage}}}

\newcommand{\maxdim}[2]{\ifdim#1>#2 #1 \else #2 \fi} 

\makeatother

\makeatletter

\newcommand{\CiteLeftDelim}{(}

\newcommand{\CiteRightDelim}{)}

\newcommand{\CiteDelimsParens}{%
\renewcommand{\CiteLeftDelim}{(}%
\renewcommand{\CiteRightDelim}{)}}

\newcommand{\CiteDelimsEmpty}{%
\renewcommand{\CiteLeftDelim}{}%
\renewcommand{\CiteRightDelim}{}}

\newcommand{\CiteListSeparator}{;}
\newcommand{\CiteListSemicolon}{\renewcommand{\CiteListSeparator}{;}}
\newcommand{\CiteListComma}{\renewcommand{\CiteListSeparator}{,}}

\def\@citex[#1]#2{\if@filesw\immediate\write\@auxout{\string\citation{#2}}\fi%
\def\@citea{}\@cite{\@for\@citeb:=#2\do%
{\@citea\def\@citea{\CiteListSeparator\penalty\@m\ }\@ifundefined%
{b@\@citeb}{{\bf ?}\@warning%
{Citation `\@citeb' on page \thepage \space undefined}}%
{\csname b@\@citeb\endcsname}}}{#1}}
\let\@internalcite\cite
\def\cite{\CiteListSemicolon\CiteDelimsParens\def\citename##1{##1
	}\@internalcite}
\def\shortcite{\CiteListComma\CiteDelimsParens\def\citename##1{}\@internalcite}
\def\newcite{\CiteListComma\CiteDelimsParens\leavevmode\def\citename##1{{##1}
	(}\@internalciteb}
\def\ecite{\CiteListSemicolon\CiteDelimsEmpty\def\citename##1{##1
	}\@internalcite}
\def\eshortcite{\CiteListComma\CiteDelimsEmpty\def\citename##1{}\@internalcite}

\def\@citexb[#1]#2{\if@filesw\immediate\write\@auxout{\string\citation{#2}}\fi
\def\@citea{}\@newcite{\@for\@citeb:=#2\do%
{\@citea\def\@citea{\CiteListSeparator\penalty\@m\ }\@ifundefined%
{b@\@citeb}{{\bf ?}\@warning%
{Citation `\@citeb' on page \thepage \space undefined}}
\hbox{\csname b@\@citeb\endcsname}}}{#1}}%
\def\@internalciteb{\@ifnextchar [{\@tempswatrue\@citexb}
	{\@tempswafalse\@citexb[]}}

\def\@newcite#1#2{{#1\if@tempswa\CiteListSeparator #2\fi)}}

\def\@biblabel#1{\def\citename##1{##1}[#1]\hfill}

\def\@cite#1#2{\CiteLeftDelim{#1\if@tempswa\CiteListSeparator
	#2\fi}\CiteRightDelim}

\def\thebibliography#1{\section*{Bibliography\@mkboth
 {Bibliography}{Bibliography}}\list
 {}{\setlength{\labelwidth}{0pt}\setlength{\leftmargin}{20pt}
 \setlength{\itemindent}{-20pt}}
 \def\newblock{\hskip .11em plus .33em minus -.07em}
 \sloppy\clubpenalty4000\widowpenalty4000
 \sfcode`\.=1000\relax}

\def\@lbibitem[#1]#2{\item[]\if@filesw
      { \def\protect##1{\string ##1\space}\immediate
        \write\@auxout{\string\bibcite{#2}{#1}}\fi\ignorespaces}}

\def\@bibitem#1{\item\if@filesw \immediate\write\@auxout
       {\string\bibcite{#1}{\the\c@enumi}}\fi\ignorespaces}

\makeatother

\newtheorem{example}{}    

\ifNFSS

\else

\fi

\newlength{\ExampleWidth}

\newlength{\FatExampleWidth}

\newcommand{\startxlh}[1] 
 {\begin{example}
  \label{#1}
  \ \
  \rm
  \begin{minipage}[t]{\ExampleWidth}\rm}

\newcommand{\startxlv}[1] 
 {\begin{example}
  \label{#1}
  \rm
  \begin{minipage}[t]{\FatExampleWidth\rm}}

\newcommand{\stopx} 
 {\end{minipage}
  \end{example}}

\newlength{\RevealingThickness}
\newcommand{\mysection}[1]{

\section{#1}

}

\newcommand{\mysubsection}[1]{

\subsection{#1}

}

\title{\vspace{-0.5in}\LARGE\bf
A Psycholinguistically Motivated Parser for CCG
}
\author{
Michael Niv\thanks{The research reported here was conducted as part of my
Ph.D. thesis work at the University of Pennsylvania and supported by the
following grants: {\sc DARPA} N00014-90-J-1863, ARO DAAL03-89-C-0031, NSF IRI
90-16592, Ben Franklin 91S.3078C-1. Preparation of this paper was supported by
a postdoctoral fellowship at the Technion in Israel.  I am grateful to Mark
Hepple, Mitch Marcus, Mark Steedman, Val Tannen, and Henry Thompson for
helpful suggestions, and to Jeff Siskind for help with typesetting CCG
derivations.  Any errors are my own.}\\
Technion -- Israel Institute of Technology\\
Haifa, Israel\\
Internet:  niv@linc.cis.upenn.edu}

\begin{document}

\maketitle
\vspace{-0.5in}
\begin{abstract}

Considering the speed in which humans resolve syntactic ambiguity, and the
overwhelming evidence that syntactic ambiguity is resolved through selection
of the analysis whose interpretation is the most `sensible', one comes to the
conclusion that interpretation, hence parsing take place incrementally, just
about every word.  Considerations of parsimony in the theory of the syntactic
processor lead one to explore the simplest of parsers: one which represents
only analyses as defined by the grammar and no other information.

Toward this aim of a simple, incremental parser I explore the proposal
that the competence grammar is a Combinatory Categorial Grammar (CCG).  I
address the problem of the proliferating analyses that stem from CCG's
associativity of derivation. My solution involves maintaining only the
maximally incremental analysis and, when necessary, computing the maximally
right-branching analysis.  I use results from the study of rewrite systems to
show that this computation is efficient.

\end{abstract}

\mysection{Introduction}

The aim of this paper is to work towards a computational model of how humans
syntactically process the language that they hear and read.  The endpoint of
this enterprise is a precise characterization of the process that humans
follow, getting details such as timing and garden pathing exactly right.

\mysubsection{Ambiguity Resolution}
\label{parallel}

\vfill

Recently, a great deal of evidence has accumulated that humans resolve
syntactic ambiguity by considering the meaning of the available analyses and
selecting the `best' one.  Various criteria for goodness of meaning have been
advanced in the psycholinguistic literature: e.g.\ thematic compatibility and
lexical selection \cite{TrueswellTanenhaus94}, discourse felicity of definite
expressions
\cite{AltmannGarnhamHenstra94},
temporal coherence in discourse \cite{TrueswellTanenhaus91},
 grammatical function {\em vis a vis\/} given/new status \cite{Niv93},
and general world-knowledge
\cite{KawamotoFarrar93}.

\vfill

Many of the works cited above consider the timing of the ambiguity resolution
decision.
The evidence is overwhelming that ambiguity is resolved within a word or two
of the arrival of disambiguating information --- that is, when there is a
meaning-based criterion which militates toward one or another syntactically
available analysis, that analysis is selected.  Should the other analysis turn
out to be the ultimately correct analysis, a garden path will result.  Given
that the various analyses available are {\em compared\/} on various criteria
of sensibleness, it follows that these analyses are {\em constructed and
maintained in parallel\/} until disambiguating information arrives.
Indeed, there is psycholinguistic
evidence that the processor maintains the various analyses in parallel
\cite{NicolPickering93,MacDonaldJustCarpenter92}.

\vfill

Our parser, therefore, must be able to build and maintain analyses in parallel.
It must also extract from the developing parse in a {\em prompt\/} fashion all
of the semantically relevant syntactic commitments (e.g.\ predicate-argument
relations) in order to allow the interpretation module that it feeds to make
accurate evaluations of the meaning. Recovery from garden paths is not
addressed in this paper.

\vfill

\mysubsection{Parser and Grammar}

Let us adopt the widely held position that humans posses a representation of
grammatical competence which is independent of any process (e.g.\ production,
perception, acquisition) that uses it.  Steedman \shortcite{GandP} argues that
if two theories of the grammar and processor package have identical empirical
coverage, but one has a more complex parser, then the other is preferred.
This preference
is not just on philosophical grounds of cleanliness of one's theories, but
stems from consideration of the evolution of the human linguistic capacity: A
theory whose grammar requires a complex parser in order to be of any use would
entail a more complex or less likely evolutionary path which the parser and
grammar took together than would a theory whose grammar requires little
specialized apparatus by way of a parser, and could thus have evolved
gradually.

So what is the simplest parser one can construct?  In other words, what is the
minimal addition of computational apparatus to the competence grammar
necessary to make it parse?  From the argument in \refsec{parallel}, this
addition must include a mechanism for maintaining analyses in parallel.
Minimally, nothing else is necessary --- the data structure which resides in
each parallel slot in the parser is a direct representation of an analysis as
defined by the competence machinery.

Suppose the grammatical competence is one that always divides an English
clause into a subject and a predicate (VP henceforth).  Suppose also that the
primary operations of the grammar are putting constituents together.  Could
the minimal parser for such a grammar account for the minimal pair in
\ref{FlowersSent}?

\startxlh{FlowersSent}
a. The doctor sent for the patient arrived.\\
b. The flowers sent for the patient arrived.
\stopx

\ref{FlowersSent}a is a garden path.  In \ref{FlowersSent}b the garden
path is avoided because flowers are not good senders.  The difference between
\ref{FlowersSent}a and b indicates that well before the word `arrived' is
encountered, the processor has already resolved the ambiguity introduced by
the word `sent'.  That is, in the main-verb analysis of `sent', the interpreter
is aware of the relation between the subject the verb {\em before\/} the end
of the VP.  But the minimal parser cannot put the subject together with `sent'
or `sent for the' because the latter are not a complete VP!

There are two possible solutions to this problem, each relaxes one of the two
suppositions above: \newcite{GandP} argues for a grammatical theory (CCG)
which does not always make the subject-predicate juncture the primary division
point of a clause. \newcite{ShieberJohnson93} on the other hand, argue that
there is no need to assume that a constituent has to be complete before it is
combined with its sister(s).  At this time, neither approach is sufficiently
developed to be evaluable (e.g.\ they both lack broad coverage grammar) so
either one is viable.  In this paper, I develop the first.

\mysection{Preliminaries}

CCG is a lexicalized grammar formalism --- a lexicon assigns each word
to one or more grammatical categories.  Adjacent constituents can combine by
one of a small number of combinatory rules.  The universe of grammatical
categories contains a collection of basic categories (e.g.\ atomic symbols such
as n, np, s, etc.\ or Prolog terms such np(3,sg)) and is closed
under the category-forming connectives / and \?.
Intuitively a constituent of category X/Y (resp.\ X\?Y) is something of
category X which is missing something of category Y to its right (resp.\
left).
The combinatory rules are listed\footnote{Two common combinatory
rules, type-raising and substitution are not listed here.  The substitution
rule \cite{CGPG} is orthogonal to the present discussion and can be added
without modification.  The rule for type-raising (see e.g.\ \ecite{Dowty88})
can cause difficulties for the parsing scheme advocated here \cite{Hepple87}
and is therefore assumed to apply in the lexicon.  So a proper name, for
example, would be have two categories: np and s/(s\?np).} in
\reftable{CombinatoryRules}.
They formalize this intuition.
A combinatory rule may be qualified with a predicate over the variables X, Y,
and Z\SUB{1}\ldots Z\SUB{n}.

\newcommand{\rulepair}[2]{\rulepairHelper{\underline{#1}}{#2}}
\newcommand{\rulepairHelper}[2]{
X/Y & Y#1 & \GoesTo & X#1 & \forw #2 & Y#1 & X\?Y & \GoesTo & X#1 & \back #2 }
\newcommand{\MovedLeft}[1]{\(\!\!\!\!\! \mbox{#1}\)}
\begin{table*}[t]
\centerline{
\begin{tabular}{|lllll|lllll|}
\hline
\multicolumn{4}{|c}{\hspace{\fill} Forward Combination \hspace{\fill} rule}
	& \MovedLeft{name} &
\multicolumn{4}{c}{\hspace{\fill} Backward Combination \hspace{\fill} rule}
	& \MovedLeft{name} \\
\hline
\rulepair{}{0} \\
\rulepair{\up Z}{1} \\
\rulepair{\up Z\SUB{1}\up Z\SUB{2}}{2} \\[-1mm]
 & & \multicolumn{1}{c}{$\vdots$} & & & & & \multicolumn{1}{c}{$\vdots$} &
	& \\[-1mm]
\rulepair{\up Z\SUB{1}\ldots\up Z\SUB{n}}{n} \\[-1mm]
 & & \multicolumn{1}{c}{$\vdots$} & & & & & \multicolumn{1}{c}{$\vdots$} &
	& \\
\hline
\end{tabular}}

\vspace{1ex}

\centerline{\up Z stands for either /Z or \?Z.  Underlined regions in a
rule must match.}
\caption{The combinatory rules}
\label{CombinatoryRules}
\end{table*}

A {\em derivation\/} is a binary tree whose leaves are each a single-word
constituent, and whose internal nodes are each
a constituent which is
derived from its children by an application of one of the combinatory rules.
A string $w$ is grammatical just in case there exists a derivation
whose frontier is $w$. I equivocate between a derivation and the constituent
at its root.  An {\em analysis\/} of a string $w$ is a sequence of derivations
such that the concatenation of their frontiers is $w$.

\mysection{The Simplest Parser}

Let us consider the simplest conceivable parser.  Its specification is ``find
all analyses of the string so far.''  It has a collection of slots for
maintaining one analysis each, in parallel.  Each slot maintains an analysis
of the string seen so far --- a sequence of one or more derivations.  The
parser has two operations, as shown in \reffig{ScanCombine}.

\begin{figure}
\begin{tabbing}
$\bullet$ {\bf scan}\\
\ \ \ \ \= get the next word from the input stream\\
\> for each analysis $a$ in the parser's memory\\
\> \ \ \ \= empty the slot containing $a$\\
\>       \> for each lexical entry $e$ of the word\\
\>       \> \ \ \ \= make a copy $a'$ of $a$\\
\>                \> add the leaf derivation $e$ to the right of $a'$\\
\>	           \> add $a'$ as a new analysis\\[1ex]
$\bullet$ {\bf combine}\\
\> for each analysis $a$ in the parser's memory\\
\> \ \ \ \= if $a$ contains more than one constituent\\
\>       \> and some rule can combine the rightmost\\
\>       \> \ \  two constituents in $a$\\
\>       \> then \= make a copy $a'$ of $a$\\
\>       \>      \> replace the two constituents of $a'$ by\\
\>       \>      \> \ \  their combination\\
\>       \>      \> add $a'$ as a new analysis
\end{tabbing}
\caption{Parser operations}
\label{ScanCombine}
\end{figure}

This parser succeeds in constructing the incremental analysis
\ref{MoreFlowers} necessary for solving the problem in \ref{FlowersSent}.

\startxlh{MoreFlowers}
\BN{s/pp}
   {\BN{s/(s\?np)}
       {\LN{the}{s/(s\?np)/n}}
       {\LN{flowers}{n}}
       {\forw 0}}
   {\LN{sent}{s\?np/pp}}
   {\forw 1}
\stopx

But this parser is just an unconstrained shift-reduce parser that simulates
non-determinism via parallelism.  It suffers from a standard problem of simple
bottom-up parsers: it can only know when a certain substring has a derivation,
but in case a substring does not have a derivation, the parser cannot yet know
whether or not a larger string containing the substring will have a derivation.
This means that when faced with a string such as

\startxlh{Insults}
The insults the new students shouted at the teacher were appalling.
\stopx

\noindent the parser will note the noun-verb ambiguity of `insults', but will
be unable to use the information that `insults' is preceded by a determiner to
rule out the verb analysis in a timely fashion.  It would only notice the
difficulty with the verb analysis after it had come to the end of the string
and failed to find a derivation for it.  This delay in ruling out doomed
analyses means that the parser and the interpreter are burdened with a quickly
proliferating collection of irrelevant analyses.

Standard solution to this problem (e.g.\ Earley's \eshortcite{Earley70}
parser; LR parsing, \ecite{AhoJohnson74}) consider global properties of the
competence grammar to infer that no grammatical string will begin with a
determiner followed by a verb.  These solutions exact a cost in complicating
the design of the parser: new data structures such as dotted rules or an LR
table must be added to the parser.  The parser is no longer a generic search
algorithm for the competence grammar.  Given the flexibility of CCG
derivations, one may consider imposing a very simple constraint on the parser:
every prefix of a grammatical string must have a derivation.  But such a move
it too heavy-handed.  Indeed CCG often gives left-branching derivations, but
it is not purely left-branching.  For example, the derivation of a
WH-dependency requires leaving the WH-filler constituent uncombined until the
entire gap-containing constituent is completed, as in
\ref{BaconSpamBeansSpamEggsSpamAndBeans}.

\startxlv{BaconSpamBeansSpamEggsSpamAndBeans}
{\vspace{.1ex}

\BN{q}
   {\BN{q/(s/np)}
       {\LN{whose}{q/(s/np)/n}\ }
       {\hspace{-0.3em}\LN{cat}{n}}
       {\forw 0}\hspace{.4em}}
   {\BN{s/np}
       {\BN{s/(s\?np)}
           {\LN{did}{s/s}\ }
	   {\LN{Fred}{s/(s\?np)}}
	   {\forw 1}\ }
       {\LN{find}{s\?np/np}}
       {\forw 1}}
   {\forw 0}
}
\stopx

\mysection{The Viable Analysis Criterion}
\label{sec:Viable}

Given the desideratum to minimize the complexity of the biologically specified
parser, I propose that the human parser is indeed as simple as the
scan-combine algorithm presented above, and that the ability to rule out
analyses such as determiner+verb is not innate, but is an acquired skill.
This `skill' is implemented as a criterion which an analysis must meet in
order to survive.  An infant starts out with this criterion completely
permissive.  Consequently it cannot process any utterances longer than a few
words without requiring excessively many parser slots.  But as the infant
observes the various analyses in the parser memory and tracks their respective
outcomes, it notices that certain sequences of categories {\em never\/} lead to
a grammatical overall analysis.  After observing an analysis failing a certain
number of times and never succeeding, the child concludes that it is not a
viable analysis and learns to discard it.  The more spurious analyses are
discarded, the better able the child is to cope with longer strings.

The collection of analyses that are maintained by the parser is therefore
filtered by two independent processes: The Viable Analysis Criterion is a
purely syntactic filter which rules out analyses independently of ambiguity.
The interpreter considers the semantic information of the remaining analyses
in parallel and occasionally deems certain analyses more sensible than
their competitors, and discards the latter.

Given that English sentences rarely require more than two or three CCG
constituents at any point in their parse, and given the limited range of
categories that arise in English, the problem of learning the viable analysis
criterion from data promises to be comparable to other n-gram learning tasks.
The empirical validation of this proposal awaits the availability of a broad
coverage CCG for English, and other languages.\footnote{In addition to the
category-ambiguity problem in \ref{Insults}, the viable analysis criterion
solves other problems, analogous to shift-reduce ambiguities, which are
omitted here for reasons of space.  The interested reader is referred to
\newcite{NivThesis} for a comprehensive discussion and an implementation of
the parser proposed here.}

\mysection{CCG and flexible derivation}

\mysubsection{The Problem}

CCG's distinguishing characteristic is its derivational flexibility --- the
fact that one string is potentially assigned many truth-conditionally
equivalent analyses.  This feature is crucial to the present approach of
incremental parsing (as well as for a range of grammatical phenomena, see
e.g.\ Steedman \eshortcite{CGPG,GandP}; \ecite{Dowty88}).  But the additional
ambiguity, sometimes referred to as `spurious', is also a source of difficulty
for parsing.  For example, the truth-conditionally unambiguous string `John
was thinking that Bill had left' has CCG derivations corresponding to each of
the 132 different binary trees possible for seven leaves.  The fact that this
sentence makes no unusual demands on humans makes it clear that its
exponentially proliferating ambiguous analyses are pruned somehow.  The
interpreter, which can resolve many kinds of ambiguity, cannot be used to for
this task: it has no visible basis for determining, for example, that the
single-constituent analysis `John was thinking' somehow makes more sense (in
CCG) than the two-constituent analysis `John'$+$`was thinking'.

Note that the maximally left-branching derivation is the one which most
promptly identifies syntactic relations, and is thus the preferred derivation.
It is possible to extend the viable analysis criterion to encompass this
consideration of efficiency as well.  The infant learns that it is usually
most efficient to combine whenever possible, and to discard an analysis in
which a combination is possible, but not taken.\footnote{Discussion of the
consequences of this move on the processing of picture noun extractions and
ambiguity-related filled-gap effects is omitted for lack of space.  See
\newcite{NivThesis}.}.

While this left-branching criterion eliminates the inefficiency due to
flexibility of derivation, it gives rise to difficulties with \ref{madly}.

\startxlh{madly}
\BN[0in]{}
   {\BN[0in]{}
       {\LN{John}{s/vp}}
       {\LN{loves}{vp/np}}
       {}}
   {\BN[0in]{}
       {\LN{Mary}{np}}
       {\LN{madly}{vp\?vp}}
       {}}
   {}
\stopx

In \ref{madly}, it is precisely the {\em non-}left-branching derivation of
`John loves Mary' which is necessary in order to make the VP constituent
available for combination with the adverb. (See \ecite{lazy}.)

\mysubsection{Previous Approaches}

Following up on the work of \newcite{Lambek58} who proposed that the process
of deriving the grammaticality of a string of categories be viewed as a proof,
there have been quite a few proposals put forth for computing only normal
forms of derivations or proofs (\ecite{Koenig89,HepMor89,Hepple91}; {\em inter
alia}).  The basic idea with all of these works is to define `normal forms'
--- distinguished members of each equivalence class of derivations, and to
require the parser to search this smaller space of possible derivations.
But none of the proposed methods result in parsing systems which
proceed incrementally through the string.\footnote{In the case of Hepple's
\shortcite{Hepple91} proposal, a left-branching normal form is indeed
computed. But its computation must be delayed for some words, so it does not
provide the interpreter with timely information about the incoming string.}

\newcite{Karttunen89} and others have proposed chart-based
parsers which directly address the derivational ambiguity problem.  For the
present purpose, the principal feature of chart parsing --- the factoring out
of constituents from analyses --- turns out to create an encumberance: The
interpreter cannot compare constituents, or arcs, for the purposes of
ambiguity resolution. It must compare analyses of the entire prefix so far,
which are awkward to compute from the developing chart.

\newcite{lazy} propose the following strategy:
(which can be taken out of the chart-parsing context of their paper)
construct only maximally left-branching derivations, but allow a limited form
of backtracking when a locally non-left-branching derivation turns out to have
been necessary.  For example, when parsing \ref{madly}, Pareschi and
Steedman's algorithm constructs the left branching analysis for `John loves
Mary'.  When it encounters `madly', it applies \forw 0 in reverse to {\em
solve} for the hidden VP constituent `loves Mary' by {\em subtracting\/} the
s/vp category `John' from the s category `John loves Mary':

\newcommand{\DoMadlyR}{ 
\SetPivot{\LN{John}{s/vp}}
\UN{s}
   {\BNp{vp}
        {\UNp[\RevealingThickness]
             {vp}
             {\BN{s}
                 {\BN{s/np}
                     {\UsePivot}
                     {\LN{loves}{vp/np}}
                     {\forw 1}}
                 {\LN{Mary}{np}}
                 {\forw 0}}
             {reveal \forw 0}}
        {\LN{madly}{vp\?vp}}
        {\back 0}}
    {\forw 0}}
\startxlh{madly-r}
\DoMadlyR
\stopx

The idea with this `revealing' operation is to exploit the fact that the rules
\forw$n$ and \back$n$, when viewed as three-place relations, are functional in
all three arguments.  That is, knowledge any two of \{left constituent,
right constituent, result\}, uniquely determines the third.
There are many problems with the completeness and soundness Pareschi and
Steedman's proposal \cite{Hepple87,NivThesis}.  For example, in \ref{bcbc},
the category b\?c cannot be revealed after it had participated in two
combinations of mixed direction: \back 0 and \forw 0.

\startxlh{bcbc}
\BN[0ex]
   {}
   {\UN[0ex]
       {}
       {\BN{a}
           {\NLN{a/b}}
           {\BN{b}
               {\BN{d}{\NLN{c}}{\NLN{d\?c}}{\back 0}}
               {\NLN{b\?d}}
               {\back 0}}
           {\forw 0}}
       {stuck}}
   {\NLN{b\?c\?(b\?c)}}
   {}
\stopx

\mysection{A Proposal}

Pareschi and Steedman's idea of lazy parsing is very attractive in the present
setting. I propose to replace their unification-based revealing operation with
a normal-form based manipulation of the derivation history.  The idea is to
construct and maintain the maximally incremental, left-branching derivations.
(see section~\ref{sec:Viable}.)  When a constituent such as the VP `loves
Mary' in \ref{madly} may be necessary, e.g.\ whenever the right-most
constituent in an analysis is of the form X\?Y, the next-to-right-most
derivation is rewritten to its equivalent right-branching derivation by
repeated application the local transformations $\arr$ defined in
\ref{drs-forw} and \ref{drs-back}.  The right frontier of the rewritten
derivation now provides all the grammatically possible attachment sites.

\newcommand{\SEMA}[1]{} 
\newcommand{\whys}{\up Y\SUB{1} $\!\!\cdots\!$ \up Y\SUB{m-1}}
\newcommand{\zees}{\up Z\SUB{1} $\!\!\cdots\!$ \up Z\SUB{n}}
\newcommand{\dees}[2]{d\SUB{#1} $\cdots$ d\SUB{#2}}
\startxlv{drs-forw}
\mbox{\BN{W $\!$\whys \zees\SEMA{ : \cB\SUP{n} (\cB\SUP{m} a b) c}}
         {\BN{W$\!$ \whys /Y\SUB{m}\SEMA{ : \cB\SUP{m} a b}}
             {\NLN{W/X\SEMA{ : a}}}
             {$\!\!\!$\NLN{X $\!$\whys /Y\SUB{m}\SEMA{ : b}}}
	     {\forw m}}
         {$\!\!\!$\NLN{Y\SUB{m} \zees\SEMA{ : c}}}
         {\forw n}} \\[3mm]
\centerline{$\arr$}\\[-2mm]
\mbox{\BN{W $\!$\whys \zees\SEMA{ : \cB\SUP{m+n-1} a (\cB\SUP{n} b c)}}
         {\NLN{W/X\SEMA{ : a}}}
         {\BN{X $\!$\whys \zees\SEMA{ : \cB\SUP{n} b c}}
             {$\!\!\!$\NLN{X $\!$\whys /Y\SUB{m}\SEMA{ : b}}}
	     {$\!\!\!$\NLN{Y\SUB{m} \zees\SEMA{ : c}}}
             {\forw n}}
         {\forw m+n-1}}
\stopx

\startxlv{drs-back}
\mbox{\BN{W $\!$\whys \zees\SEMA{ : \cB\SUP{m+n-1} a (\cB\SUP{n} b c)}}
         {\BN{X $\!$\whys \zees\SEMA{ : \cB\SUP{n} b c}}
             {\NLN{Y\SUB{m} \zees\SEMA{ : c}}}
             {$\!\!\!$\NLN{X $\!$\whys \?Y\SUB{m}\SEMA{ : b}}}
	     {\back n}}
	 {$\!\!\!$\NLN{W\?X\SEMA{ : a}}}
         {\back m+n-1}}\\[3mm]
\centerline{$\arr$}\\[-2mm]
\mbox{\BN{W $\!$\whys \zees\SEMA{ : \cB\SUP{n} (\cB\SUP{m} a b) c}}
         {\NLN{Y\SUB{m} \zees\SEMA{ : c}}}
         {\BN{W $\!$\whys \?Y\SUB{m}\SEMA{ : \cB\SUP{m} a b}}
             {$\!$\NLN{X $\!$\whys \?Y\SUB{m}\SEMA{ : b}}}
             {$\!$\NLN{W\?X\SEMA{ : a}}}
             {\back m}}
        {\back n}}
\stopx

Results from the study of rewrite systems (see \newcite{Klop92} for an
overview) help determine the computational complexity of this operation:

\mysubsection{A Rewrite System for Derivations}

\newcommand{\NN}{\mbox{\#}}
\newcommand{\LeftChild}{\mbox{\(\lambda\)}}
\newcommand{\RightChild}{\mbox{\(\rho\)}}

If $x$ is a node in a binary tree let
\LeftChild$(x)$ (resp.\ \RightChild$(x)$) refer to its left (right) child.

Any subtree of a derivation which matches the left-hand-side of either
\ref{drs-forw} or \ref{drs-back} is called a {\em redex}.  The result of
replacing a redex by the corresponding right-hand-side of a rule is called the
{\em contractum}.  A derivation is in {\em normal form (NF)\/} if it contains
no redexes. In the following I use the symbol $\arr$ to also stand for the
relation over pairs of derivations such that the second is derived from the
first by one application of $\arr$.  Let $\larr$ be the converse of $\arr$.
Let $\longleftrightarrow$ be $ \arr \cup \larr$.  Let $\darr$ be the reflexive
transitive closure of $\arr$ and similarly, $\invdarr$ the reflexive
transitive closure of $\larr$, and $\drpl$ the reflexive transitive closure of
$\longleftrightarrow$.  Note that $\drpl$ is an equivalence relation.

A rewrite system is {\em strongly normalizing} (SN) iff every sequence of
applications of $\arr$ is finite.

\theorem{\label{drs-sn}$\arr$ is SN\footnote{\newcite{HepMor89} Proved SN for
a slight variant of $\arr$.  The present proof provides a tighter score
function, see lemma~\ref{drs-upper-bound} below.}}

\newcommand{\wate}{\NN}
\newcommand{\skore}{\mbox{{$\sigma$}}}
\newcommand{\kost}{\mbox{{$\kappa$}}}
\newcommand{\drm}{\mbox{{$d_{\mbox{rm}}$}}}

\pf{
Every derivation with $n$ internal nodes is assigned a positive integer
score.  An application of $\arr$ is guaranteed to yield a derivation with a
lower score.  This is done by defining functions \wate\ and \skore\ for each
node of the derivation as follows:

\[
\begin{array}{lll}
\wate(x) & \!\!\!=\!\!\! & \!\!\left\{\!\!\!
\begin{array}{ll}
0 \mbox{\ if $x$ is a leaf node}&\\
 1 + \wate(\mbox{\LeftChild($x$)}) + \wate(\mbox{\RightChild($x$)}) &
\mbox{otherwise}
\end{array}\right. \\ 
\mbox{\rule{0pt}{.1mm}} & & \\
\skore(x) & \!\!\!=\!\!\! & \!\!\left\{\!\!\!
\begin{array}{ll}
0 \mbox{\ if $x$ is a leaf node} &\\
\skore(\mbox{\LeftChild($x$)})+\skore(\mbox
	{\RightChild($x$)})+\wate(\mbox{\LeftChild($x$)})&
			\mbox{\hspace{-0.5em}otherwise}
\end{array}\right. 
\end{array}
\]

Each application of $\arr$ decreases \skore, the score of the derivation.
This follows from the monotonic dependency of the score of the root of the
derivation upon the scores of each sub-derivation, and from the fact that
locally, the score of a redex decreases when $\arr$ is applied:
In \reffig{score}, a derivation is depicted schematically with a redex whose
sub-constituents are named a, b, and c.
Applying $\arr$ reduces
\skore(e), hence the score of the whole derivation.

\begin{figure}
\centerline{\psfig{figure=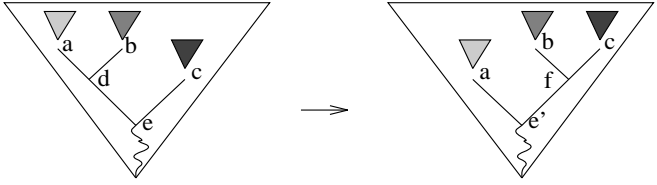,silent=y,width=3.4in}}
\caption{Schema for one redex in DRS}
\label{score}
\end{figure}

\noindent in redex:
\begin{eqnarray*}
\wate(d)  & = & \wate(a) + \wate(b) + 1 \\
\skore(d) & = & \skore(a) + \skore(b) + \wate(a) \\
\skore(e) & = & \skore(d) + \skore(c) + \wate(d) \\
          & = & \skore(a) + \skore(b) + \skore(c) + \wate(b) + 2\cdot\wate(a)
                             + 1
\end{eqnarray*}

\noindent in contractum:
\begin{eqnarray*}
\skore(f)  & = & \skore(b) + \skore(c) + \wate(b) \\
\skore(e') & = & \skore(a) + \skore(f) + \wate(a) \\
           & = & \skore(a) + \skore(b) + \skore(c) + \wate(b) + \wate(a) \\
	& < & \skore(a) +
\skore(b) + \skore(c) + \wate(b) + 2\cdot\wate(a) + 1
\end{eqnarray*}} 

\noindent Observe that $\wate(x)$ is the number of internal nodes in $x$.

\lemma{\label{drs-upper-bound} Given a derivation $x$, let $n=\wate x$.
Every sequence of applications of $\arr$ is of length at most
$n(n-1)/2$.\footnote{\newcite{NivDRS} shows by example
that this bound is tight.}}

\pf{
By induction on $n$:\\
Base case: $n=1$; 0 applications are necessary.\\
Induction: Suppose true for all derivations of fewer than $n$ internal nodes.
Let $m=\NN\LeftChild(x)$. So $0 \leq m \leq n-1$ and
$\NN\RightChild(x)=n-m-1$.
{\newcommand{\Movit}{\hspace{5mm}}
\[ \begin{array}{l}
\skore(x)-n(n-1)/2 = \hspace{\fill}\\[.4ex]
\Movit = \skore(\LeftChild(x))+\skore(\RightChild(x))+\wate(\LeftChild(x))-
				n(n-1)/2\\[.4ex]
\Movit \leq \frac{m(m-1)}{2} + \frac{(n-m-1)(n-m-2)}{2} + m -
		\frac{n(n-1)}{2}\\[.4ex]
\Movit = (m+1)(m-(n-1))\\[.4ex]
\Movit \leq 0 \mbox{\hspace{3em} recalling that $0 \leq m \leq n-1$}
\end{array} \]
}
}

So far I have shown that every sequence of applications of $\arr$ is not very
long: at most quadratic in the size of the derivation.  I now show that when
there is a choice of redex, it makes no difference which redex one picks.
That is, all redex selection strategies result in the same normal form.

A rewrite system is {\em Church-Rosser (CR)} just in case
\[ \forall x,y . (x \drpl y \implies \exists z . (x \darr z \And y \darr z)) \]

A rewrite system is {\em Weakly Church-Rosser (WCR)} just in case \[ \forall
x,y,w . (w \arr x \And w \arr y) \implies \exists z . (x \darr z \And y \darr
z) \]

\lemma{\label{l:wcr} $\arr$ is WCR.}

\begin{figure}
\centerline{\psfig{figure=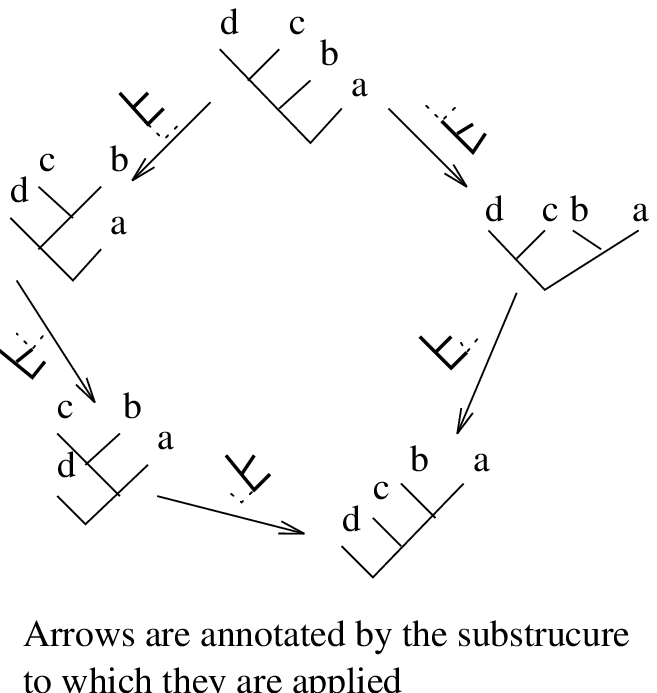,silent=y}}
\caption{Why $\arr$ is weakly Church-Rosser}
\label{wcr}
\end{figure}

\vbox{
\pf{Let $w$ be a derivation with two distinct redexes $x$ and $y$, yielding
the two distinct derivations $w'$ and $w''$ respectively.  There are a few
possibilities:
\kases
\kase{1} $x$ and $y$ share no internal nodes.  There are three
subcases: $x$ dominates $y$ (includes $y$ as a subconstituent), $x$ is
dominated by $y$, or $x$ and $y$ are incomparable with respect to dominance.
Either way, it is clear that the order of application of $\arr$ makes no
difference.
\kase{2} $x$ and $y$ share some internal node.  Without
loss of generality, $y$ does not dominate $x$.
There exists a derivation $z$ such that $w' \darr z \And w'' \darr z$.
This is depicted in \reffig{wcr}.
(Note that all three internal nodes in \reffig{wcr} are of the same rule
direction, either \forw\ or \back.)
\Endkases
}
}

\lemma{\label{newman} (Newman)
$\mbox{WCR} \And \mbox{SN} \implies \mbox{CR}$.}

\theorem { $\arr$ is CR. }

\pf{From theorem \ref{drs-sn} and lemmas \ref{l:wcr} and \ref{newman}.}

Therefore any maximal sequence of applications of \mbox{$\arr$} will lead to
the normal form\footnote{Assuming, as is the case with extant CCG accounts,
that constraints on the applicability of the combinatory rules do not present
significant roadblocks to the derivation rewrite process.}.  We are free to
select the most efficient redex selection scheme.  From
lemma~\ref{drs-upper-bound} the worst case is quadratic. \newcite{NivDRS}
shows that the optimal strategy, of applying $\arr$ closest as possible to the
root, yields $\arr$ applications sequences of at most $n$ steps.

Note that all that was said in this section generalizes beyond CCG derivations
to any associative algebra.

\mysubsection{Discussion}

Given the rightmost subconstituent recovered using the normal form technique
above, how should parsing proceed?  Obviously, if the
leftward looking
category which precipitated the normal form computation is a modifier, i.e.\
of the form X\?X, then it ought to be combined with the recovered constituent
in a form analogous to Chomsky adjunction.  But what if this category is not
of the form X\?X?  For example, should the parser compute the reanalysis in
\ref{nonmonotonic-reanalysis}?

\startxlh{nonmonotonic-reanalysis}
\BN[0pt]
   {}
   {\BN{a/d}
       {\BN{a/c}
           {\NLN{a/b}}
           {\NLN{b/c}}
           {\forw 1}}
       {\NLN{c/d}}
       {\forw 1}}
   {\NLN{s\bkslf(a/b)\bkslf(b/d)}}
   {}\hspace{1cm}
\BN{s}
   {\NLN{a/b}}
   {\BN{s\bkslf(a/b)}
       {\BN{b/d}
           {\NLN{b/c}}
           {\NLN{c/d}}
           {\forw 1}}
       {\NLN{s\bkslf(a/b)\bkslf(b/d)}}
       {\back 0}}
   {\back 0}
\stopx

Ascribing the same non-garden-path status to the reanalysis in
\ref{nonmonotonic-reanalysis} that we do to \ref{madly-r} would
constitute a very odd move: Before reanalysis, the derivation encoded the
commitment that the /b of the first category is satisfied by the b of
the b/c in
the second category.  This commitment is undone in the reanalysis.  This is
an undesirable property to have in a computational model of parsing
commitment, as it renders certain revisions of commitments easier than others,
without any empirical justification.  Furthermore, given the possibility that
the parser change its mind about what serves as argument to what, the
interpreter must be able to cope with such non-monotonic updates to its view
of the analysis so far --- this would surely complicate the design of the
interpreter.\footnote{I am indebted to Henry Thompson for a discussion of
monotonicity.} Therefore, constituents on the right-frontier of a
right-normal-form should only combine with `endocentric' categories to their
right.  The precise definition of `endocentric' depends on the semantic
formalism used --- it certainly includes post-head modifiers, and might also
include coordination.

Stipulating that certain reanalyses are impossible immediately makes the
parser `incomplete' in the sense that it cannot find the analysis in
\ref{nonmonotonic-reanalysis}. From the current perspective of identifying
garden paths, this incompleteness is a desirable, even a necessary property.
In \ref{nonmonotonic-reanalysis}, committing to the composition of a/b and b/c
is tantamount to being led down the garden path.  In a different sense, the
current parser is complete: it finds all analyses if the Viable Analysis
Criterion and the interpreter never discard any analyses.

\mysection{Conclusion}

The current proposal shifts some of the burden traditionally associated with
the parser to other components of the human cognitive faculty: the interpreter
resolves ambiguity, and an acquired skill removes `garbage' analyses from the
parser's memory --- solving the so-called spurious ambiguity problem, as well
as effectively applying grammar-global constraints traditionally computed by
top-down techniques or grammar compilation.  The resultant parser adheres to
the desideratum that it be a generic search algorithm for the grammar
formalism, provided the definition of CCG explicitly includes the notion of
`derivation' and explicates the truth-conditional equivalence relation.  Such
inclusions have indeed been proposed \cite{GACC}.


\end{document}